\documentclass{ecai} 

\usepackage{soul}
\usepackage[hidelinks]{hyperref}

\usepackage{xcolor}

\begin{document}

\begin{frontmatter}


\paperid{2822} 


\title{Rethinking How AI Embeds and Adapts to Human Values: Challenges and Opportunities}


\author[A]{\fnms{Sz-Ting}~\snm{Tzeng}\orcid{0000-0001-9304-6566}}
\author[A]{\fnms{Frank}~\snm{Dignum}\orcid{https://orcid.org/0000-0002-5103-8127}}

\address[A]{Ume\aa University}


\begin{abstract}
The concepts of ``human-centered AI'' and ``value-based decision'' have gained significant attention in both research and industry.
However, many critical aspects remain underexplored and require further investigation. In particular, there is a need to understand how systems incorporate human values, how humans can identify these values within systems, and how to minimize the risks of harm or unintended consequences.
In this paper, we highlight the need to rethink how we frame value alignment and assert that value alignment should move beyond static and singular conceptions of values. 
We argue that AI systems should implement long-term reasoning and remain adaptable to evolving values. 
Furthermore, value alignment requires more theories to address the full spectrum of human values.
Since values often vary among individuals or groups, multi-agent systems provide the right framework for navigating pluralism, conflict, and inter-agent reasoning about values.
We identify the challenges associated with value alignment and indicate directions for advancing value alignment research.
In addition, we broadly discuss diverse perspectives of value alignment, from design methodologies to practical applications.
\end{abstract}


\end{frontmatter}
\thispagestyle{plain}
\pagestyle{plain}



\section{Introduction}
Recent advancements in general-purpose artificial intelligence (AI) have led to a deeper integration of AI agents into our daily lives and social structures.
This integration has given rise to a novel multi-agent system where humans and AI interact and collaborate in previously unseen ways.
This symbiosis reshapes our understanding of social dynamics and challenges traditional ideas of agency and decision-making in complex environments \cite{Murukannaiah2020Blue-Sky}.
Values are the driving forces that shape attitudes and guide human behaviors \cite{Schwartz2012overview}. 
Values are not innate but are shaped by cultural and societal influences. Moreover, significant life experiences can also shape an individual's values. Thus, behavior is shaped by values, but in its turn also shapes values again.
Value importance for deliberations on behavior can vary by context. Often values only play an indirect role by shaping routines, norms, and affordances. However, they always play a role in the background, shaping the playing field on which decisions about behavior are made. Values can function at micro and macro levels \cite{Chopra2018ethics,Woodgate2022Blue-Sky,Liscio2023value}.
At the micro level, the focus is on aligning AI decisions with an individual's value importance. In contrast, at the macro level, the goal is to shape collective behaviors and establish social norms that guide the interactions between the agents in multi-agent systems \cite{Serramia2023encoding}.

While collaborative human-AI teams demonstrate significant appeal, their effectiveness remains dependent on the successful alignment of objectives across team members \cite{Mechergui2024goalAlignment}. 
The challenge of aligning AI with human values has become known as the \textit{value alignment problem} \cite{Russell2019human}.
The core aspects of addressing the value alignment problem include defining values, modeling values, integrating values into decision making, ensuring interpretability and explainability, verifying and validating the alignment, and enabling adaptability.

Despite growing interest in embedding human values into AI \cite{Russell2019human,Dignum2019responsibleAI,Gabriel2020AI}, further discussion is needed on how to effectively and ethically integrate these values into AI and real-world applications.

\subsection{Contribution}
This paper makes several key contributions to value alignment in AI, particularly through the lens of multi-agent systems and human-centered AI design. While prior work has acknowledged and provided insightful analyses of the dynamic, context-sensitive, and pluralistic nature of values \cite{Liscio2021axies,Lera2022towards}, these studies have not addressed the fundamental challenges directly. This paper takes a step further by proposing a roadmap for operationalizing these complex value dimensions in real-world contexts.
We outline the core challenges of integrating values into AI, spanning the entire process from design to application.
To advance future research for value alignment, we highlight key directions that require attention and present a set of interrelated contributions.

We articulated our central stance through three main claims:  
\begin{itemize}
    \item AI should reason about values in long-term dynamics.
    \item Value alignment requires comprehensive theoretical foundations to address the complex, interdependent, and evolving nature of human values.
    \item Multi-agent systems offer the right frame for value alignment.
\end{itemize}
To support this position, we present a structured discussion of the challenges in value alignment to highlight the reconceptualization of value alignment methodologies is essential to achieve value alignment in practical.

\paragraph{Interdisciplinary Roadmap.}
While values are inherently complex and can evolve with diverse cultural and social factors, insight from different disciplines will benefit this research. 
For example, philosophy provides foundational theories of values.
Social science contributes empirical insights into social norms and how values operate and evolve across societies.
Human-computer interaction focuses on the design of interaction between AI systems and human users, and aims to ensure these interactions reflect human values, needs, and expectations.
Cognitive science studies mental processes such as perception, attention, reasoning, and decision making.
Computer science operationalizes these insights into algorithms, models, and systems to accommodate value engineering.

\subsection{Organization}
Section~\ref{sec:problem} provides an overview of the value alignment problem.
Section~\ref{sec:claims} articulates our claims for addressing value alignment.
Section~\ref{sec:challenges} discusses challenges associated with achieving value alignment.


\section{Rethinking Values Alignment}
\label{sec:problem}

The value alignment problem refers to the challenge of ensuring that AI operates in ways that align with human values and ethical principles.
While AI can be a helpful tool for tasks with clearly and precisely defined objectives, there is a growing expectation for AI to perform beyond these limitations.
Since humans cannot always anticipate or specify every possible scenario, aligning AI with human values is essential to minimize risks of harm or unintended consequences.

Before incorporating values into systems, it is essential to define what values are \cite{Schwartz2012overview,Haidt2009aboveBelow,Rokeach1973values}. 
Some existing research in value alignment assumes that values are static or singular. For example, privacy has been approached as a static concept and represented by predefined preferences or optimization objectives \cite{Di2023paccart}.
While these approaches have driven notable technical progress, they fall short of capturing the pluralistic, dynamic, and interconnected nature of values.
Schwartz's value theory is a comprehensive and widely recognized framework for addressing the value alignment problem in AI \cite{Schwartz2012overview}.
Instead of considering values as a pure preference, Schwartz considered values as a system where certain values can support or oppose others.
For instance, values emphasizing self-enhancement, such as power and achievement, may conflict with those emphasizing self-transcendence, like benevolence and universalism.
Another important aspect of Schwartz's value theory is that values are seen as a universal set of values empirically founded worldwide. This provides a solid basis for the use of this theory in different places around the world and in different types of applications. While Schwartz's value structure may oversimplify pluralistic human values or overemphasize universality across cultures \cite{Sorensen2024value}, this theory provides a good starting point for incorporating values into systems. However, as we will discuss later, Schwartz's values are very abstract and can usually not be used to link values to behavior directly. More is needed to link them to actual behavior in a consistent way.

To embed values into systems, it is essential to extract and elicit these values from human behaviors.
This process involves engaging stakeholders and analyzing behavioral data.
Additionally, sources such as news articles or books can serve as behavioral data to provide insight into values for specific groups.
Value alignment entails value inference, which is the process of identifying and estimating value importance based on observations.
\citet{Liscio2023value} decompose value inference into value identification, value systems estimation, and value systems aggregation.
The elicitation and identification processes form the foundation for value modeling.

Systems can incorporate values into their decision-making processes through implicit or explicit methods. The implicit approach avoids directly modeling or representing values, instead relying on patterns derived from datasets \cite{Liscio2021axies,Shen2024bidirectionalAlignment}. 
Conversely, the explicit approach involves translating abstract values into structured formats that AI can process, e.g., utility function \cite{Serramia2018values,Osman2024values,Liscio2023value} or logic-based approaches \cite{Dennis2016formal}.
\citet{Osman2024values} propose a computational model of human values to provide the foundation for systems to make value-aligned decisions.
\citet{Lera2022towards} address the pluralistic value alignment problem where the AI system aligns with a group of people with different ethical principles (e.g., maximum utility and maximum fairness). 
\citet{Rodriguez2023MORL} combine Multi-Objective Reinforcement Learning and Linear Programming to build agents that are in alignment with multiple moral values.

Interpretability, explainability, verification, and validation are interconnected and critical to addressing the value alignment problem.
Interpretability refers to the user's ability to comprehend the system's output, whereas explainability refers to providing human-understandable explanations for the decisions \cite{Miller2019explanation}. 
Verification ensures that the system's implementation correctly incorporates human values, while validation ensures the decisions align with human values or societal principles.
Interpretability and explainability support verification and validation.
Verification and validation provide insights to improve interpretability and explainability.
Ensuring interpretability and explainability is essential for helping users and stakeholders understand the reasoning behind decisions and building trust in systems. 

Given the dynamic and interdependent nature of human values, AI in multi-agent systems must be able to adapt to changes in values.
Specifically, values are not static and usually evolve with social, cultural, and environmental influences over time.
Aligning an AI system with a static objective risks oversimplifying complex ethical and social considerations and may fail in dynamic environments.
For instance, when the awareness of environmental sustainability grows, AI systems may need to adapt to evolving regulations and social norms and prioritize sustainability over other values in decisions.
These limitations highlight the importance of rethinking our approach to value alignment. We argue for developing comprehensive theoretical frameworks and a richer framing of value alignment that regards values as long-term dynamics, and within the lens of multi-agent systems.


\section{Core Claims}
\label{sec:claims}
This section articulates three interrelated claims about our vision of value alignment in AI. Each claim is grounded in theoretical insight and motivated by real-world scenarios. These claims guide the position we advocate for in this paper.

\subsection{Deeper Theoretical Foundations}
No single set of values universally applied across all individuals or cultures.
Instead, individuals and groups often prioritize divergent values, which may sometimes conflict or support one another, or be challenging to compare. Even within a single community, consensus is not always guaranteed. 
The challenge of value alignment involves complex social, cultural, ethical, and psychological dimensions that extend beyond the scope of computer science alone. 
Addressing these dimensions requires the integration of interdisciplinary perspectives. 
Prior work often overlooks the complexity, interdependence, and evolution of values.
Achieving value alignment requires the development of new theories or comprehensive theoretical frameworks to accommodate the complex, interdependent, and evolving nature of human values. 

\paragraph{Motivating Examples:}
In moderating content on social media platforms, values like freedom of expression, protection from harm, cultural norms, and fairness across diverse user groups or roles often conflict. What is considered acceptable or harmful can differ significantly across cultures and groups. 
Moreover, prioritizing freedom of expression may conflict with protection from harm for one user but support for respect for another. Some may perceive a sarcastic meme as humorous, while another finds it offensive.
AI systems must navigate this dynamics and complexity while performing sensitive mediation.

\paragraph{Supporting Challenges:}
\begin{itemize}
    \item Value Elicitation and Identification: Comprehensive theoretical foundations are essential to accommodate values' interdependencies and context-dependence, and provide structured approaches to navigate conflicts.
    \item Value Integration: No consensus exists for modeling interdependencies, especially when those values are conflicting or incommensurable, or for resolving the conflicts.
    \item Verification and Validation: How does ``correct'' alignment looks like?
    \item Interpretability and Explainability: Explanations should go beyond describing the action itself and include the reasoning behind the chosen trade-off and the values considered in decision making.
    \item Value Source: Whose values should AI systems align with, and which specific values should be prioritized?
    \item Interdisciplinary Research: A coherent value framework requires insights that reach beyond computer science.
\end{itemize}


\paragraph{Position:}
Value alignment requires comprehensive theoretical foundations to accommodate the dynamism and interdependencies of values.

\subsection{Value Alignment as Long-Term Dynamics}
Values are dynamic and evolving with cultural and environmental influences. The interplay between values and human behavior is mutual, with each continuously shaping and evolving in response to the other.
In addition, short-term compromises are common in the real world, but individuals tend to realign with their values over extended periods.
Therefore, designing AI systems that align with human values requires moving beyond static preferences, fixed utility, static mappings, or short-term optimization.
These systems must be capable of interpreting, reasoning, and adapting to these dynamic standards.

\paragraph{Motivating Examples:}
A virtual assistant must consider individual values and temporal contexts. When a manager instructs an employee who values honesty and transparency to temporarily withhold certain information from a client during ongoing negotiations, maintaining the company’s negotiation position and following the manager’s instructions may lead to a short-term compromise of transparency.

\paragraph{Supporting Challenges:}
\begin{itemize}
    \item Value Elicitation and Identification: The processes of elicitation and identification must consider that values reflected in data or behavior can vary over time and in different contexts. Furthermore, the process should accommodate the possibility of short-term compromises.
    \item Value Estimation and Aggregation: Estimation and aggregation processes must consider how values shift across time and contexts.
    \item Abstract and Concrete Values: Moving from abstract values to concrete and context-specific behaviors requires dynamic interpretation.
    \item Applications: Determining which applications require value alignment and how to balance system autonomy with alignment mechanisms is critical.
\end{itemize}




\paragraph{Position:}
AI systems should be designed to reason about values within long-term dynamics.

\subsection{Multi-Agent Systems as the Right Computational Frame}
Values may differ across individuals and often conflict, which makes single-agent alignment insufficient for functioning effectively in social or multi-stakeholder settings. AI systems must navigate pluralistic, conflicting, and dynamic values across diverse individuals and groups within these inherently multi-agent environments. Framing value alignment within a multi-agent systems perspective opens the door to mechanisms for negotiating and resolving value conflicts.

\paragraph{Motivating Examples:}
In an urban planning project, diverse stakeholders include urban planners, developers, advocacy groups, and residents participating in city land-use planning. Each individual brings unique preferences, backgrounds, and often conflicting values on issues such as land use, environmental protection, economic development, and social equity. 
After an initial plan was proposed, each group shared their values and needs, engaged in structured discussions, provided feedback, and negotiated changes. 
This process involves explicit and implicit reasoning among agents about their values and the consequences of their decisions.
The urban planner then revises the plan to reflect these pluralistic values.
AI systems should model and reason about these value interactions across individuals and enable negotiation to resolve conflicts.

\paragraph{Supporting Challenges:}
\begin{itemize}
    \item Value Estimation and Aggregation: How should values be aggregated or reconciled? This is fundamentally a normative and social choice problem, not merely an optimization problem.
    \item Value Integration: Inter-agent value conflicts require mechanisms to resolve disagreements.
    \item Value Source: Whose values should be prioritized in multi-agent systems? Who should make the decisions?
    \item Applications: In contexts such as healthcare or education, value alignments can happen among stakeholders and involve multiple agents.
    \item Interdisciplinary Research: Understanding how values emerge, interact, and are negotiated within a multi-agent system requires insights that extend beyond the field of computer science.
\end{itemize}

\paragraph{Position:}
Multi-agent systems offer the right frame for modeling pluralism, conflict, and inter-agent reasoning about values.


\section{Challenges}
\label{sec:challenges}
Achieving value alignment in AI systems involves tackling a wide range of conceptual, computational, and social challenges. While these challenges are deeply interconnected, each challenge opens space for new interdisciplinary insights and research directions. 
In this section, we organize these challenges into thematic categories and highlight how addressing them can drive meaningful progress in the field.

\subsection*{Value Elicitation and Identification}
\paragraph{Opportunity.}
Social science, psychology, and philosophy researchers delve into theories of values \cite{Rokeach1973values,Schwartz2012overview},
and interdisciplinary collaboration among these fields offers opportunities for more robust value alignment by enabling richer models of value modeling and contextual interpretation.

\paragraph{Challenge.}
Critical questions arise as value alignment gains more attention: How can values be identified in real-world scenarios? Which values are contextually relevant?
Values can be extracted from direct observations to indirect resources such as books, movies, and cultural traditions.
\citet{Schwartz1994UniversalAspectsOfValues} derived values from human subjects by combining theoretical development with empirical cross-cultural analyses using the similarity structure analysis on importance ratings.
\citet{Friedman2013VSD} propose a proactive design of systems to support the values of direct and indirect stakeholders. They identify those values through conceptual, empirical, and technical investigations.
\citet{Liscio2021axies} employ a data-driven approach to identify context-specific values from textual data.

Recognizing the differences among values is a complex challenge. 
On one hand, different values can lead to the same behaviors. On the other hand, the complex interplay of opposing and supporting values can complicate the behaviors.
While values are context-dependent and multifaceted and often vary across individuals, the perceived values and their interdependencies also depend on subjective interpretation.
Short-term compromises are acceptable while maintaining core values.
All these increase the difficulty of identifying values.

\subsection*{Value Estimation and Aggregation}
\paragraph{Opportunity.}
Mechanisms based on social choice theory create opportunities for value alignment by ethically aggregating diverse values and balancing the interests of majority and minority groups. Transparent aggregation processes further support alignment by enabling stakeholders to understand how their values are weighted and reflected in collective decisions.

\paragraph{Challenge.}
When assessing the value importance, a single behavior may reflect aggregated values \cite{Ajmeri2020elessar}. 
Similarly, value aggregation can be applied to represent a group and make collective decisions \cite{Lera2022towards}.
While values are interdependent, an individual can hold values that share similar underlying or antagonistic motivations simultaneously. 
The complexity, context dependence, and potentially conflicting and interdependent nature of hold values make the assessment more difficult.
Consistently resolving and assessing conflicting values is a significant challenge.
In addition, the ambiguity, bias, misinterpretation, and incomplete information posed by humans further challenge the assessment of values.

\subsection*{Abstract and Concrete Values}
\paragraph{Opportunity.}
Interdisciplinary collaboration is crucial for establishing links between abstract values and observable behaviors. Furthermore, innovative metrics or assessment tools are necessary to evaluate how well actions align with defined values objectively.

\paragraph{Challenge.}
While researchers define and identify values through abstract concepts \cite{Rokeach1973values,Schwartz2012overview}, different people or cultures may interpret the same concepts differently. For instance, someone might defend a factory that is polluting nature in a region with the fact that it also brings jobs and prosperity. Thus arguing from the value of transcendence that it is good for the community. However, one can also argue that polluting the environment is bad for the community, and thus, from the value of transcendence, one should be against the factory.
Abstract values are meant to guide people over many different situations, times, and places. This makes them powerful, but also causes problems when creating links between abstract values and actions. Many elements of the context will modify the link and result in different influences of values on behavior. 
A concrete definition is essential to link these abstract values to observable behaviors and outcomes \cite{Osman2024values}.
However, quantifying or objectively measuring abstract values poses significant challenges.
Additionally, encoding conflicts between values is complex and difficult to normalize.
Since values can evolve with societal changes and individual experiences, a concrete value that is feasible today can become outdated in the future.

\subsection*{Value Integration}
\paragraph{Opportunity.}
Research could explore the development of adaptive systems that dynamically adjust the importance of different values in response to contexts, user input, or observed outcomes. Another promising direction is to design mechanisms that enable agents to reason about others' values, negotiate, and reach a consensus or compromise that reflects the individual or collective decision. Additionally, research can investigate methods to maintain alignment over extended periods, such as by developing utility functions or frameworks that incorporate temporal trade-offs and compensatory behaviors.

\paragraph{Challenge.}
Systems can integrate values either explicitly or implicitly.
The implicit approach identifies correlations between context and annotated values \cite{Liscio2021axies}, whereas the explicit approach incorporates values through logical representation or utility functions \cite{Dennis2016formal,Serramia2018values,Liscio2023value}.

Logical representations enable the embedding of values through specified rules.
For example, a rule stating that doctors must not share personal medical records with anyone other than the patient without consent reflects the value of patient privacy.
These rule-based representations are transparent and consistent in outputs and easy to maintain.
However, such rules may oversimplify the complexity of values.
For instance, logical representations often overlook individual differences in value importance and the evaluation of varied consequences.
Moreover, while values are often context-based, embedding values through rules necessitates specifying an exhaustive set of conditional rules.
Real-world interactions often involve competing values, and rule-based systems must predefine the priority of values to resolve the conflicts.
Maintaining and updating rules becomes challenging as systems become complex or values evolve.

Modeling values with utility functions enables systems to model trade-offs and make optimal decisions based on the defined utility.
However, the interdependence and conflict among values pose a significant challenge to defining utility functions. Similar to explicit rules, utility functions must account for all kinds of exceptions and combinations. Moreover, they should contain a temporal dimension. For example, an individual can decide to drive to work instead of biking during severe weather and argue that he will compensate for this by refraining from using the car for a weekend trip to a museum. 
In other words, it is not necessary to consistently behave in complete alignment with one's core values as long as such alignment is maintained over a long period of time.

Another important concern is whether it is necessary to model all values. 
In some cases, incorporating specific values may be more beneficial for users, especially when stakeholders aim to promote particular values.

While values are difficult to quantify, reconciling values is also challenging.
Even when two competing values have differing levels of importance, the value deemed more important does not necessarily represent the optimal solution. 
For instance, using a bike to take children to school and then continuing to work might be optimal with respect to the values of the environment. However, this decision could lead to being late for a meeting at work. 
Such delay could reduce one's motivation to support an environmental policy at the company despite the potential benefits that would far outweigh the cost of a car ride.
In addition, utility functions lack the adaptability of dynamic environments and may overlook the unmeasurable aspects.

Learning-based approaches identify patterns or values from datasets or interactions. 
However, ensuring interpretability and that the learned behaviors reflect specific values remains a significant challenge.
Additionally, the quality of datasets used for training models raises concerns. 
Models trained on biased or non-representative data may suggest unfair or ineffective decisions and may fail to generalize effectively to real-world scenarios. 

\subsection*{Verification and Validation}
\paragraph{Opportunity.}
Research can investigate adaptive frameworks that enable systems to realign their values in response to evolving social norms through continual learning or participatory design. Additionally, developing frameworks for continuous, long-term monitoring and auditing deployed systems is essential for real-world validation.

\paragraph{Challenge.}
Verification ensures that the system's \textbf{implementation} aligns accurately with human values.
This process is grounded explicitly in the clearly defined requirements.
The specifications used for verification should be detailed and specific rather than abstract or high-level.
Verification mechanisms include formal methods \cite{Sierra2021formalVA}, simulations \cite{Brown2021VAV}, and human feedback \cite{Chakraborty2024PARL}.
However, values' interdependent and context-sensitive nature makes defining concrete, verifiable specifications difficult.
This challenge is further amplified in systems built using learning-based approaches due to the non-linear and inherently non-transparent nature. 
In addition, even if system outputs resemble human outputs in terms of shapes or patterns, there is no assurance that the model meets human expectations.
Concrete specifications are necessary for verification.

Validation confirms that the system's \textbf{outputs} align with human values.
However, the process requires concrete behavioral measurements and is challenged by the interdependent and context-sensitive nature of values.
In addition, since values may vary by context, generalizing context-sensitive evaluation remains a significant challenge.

There are further issues beyond passing the verification and validation stages.
First, verification and validation are tested in predefined scenarios with predefined measurements. However, the real world involves higher uncertainty, complexity, and dynamicity.
Second, the embedded values will deviate from current values when values evolve with environmental changes.
Third, unforeseen edge cases can cause systems to deviate from their intended alignment.
Fourth, even if the system passes the verification process through simulations, the real-world validation may require several years.

\subsection*{Interpretability and Explainability}
\paragraph{Opportunity.}
Research can explore tailoring explanations to individuals' values and contexts.

\paragraph{Challenge.}
Decisions should be understandable and accountable to humans to build trust in systems.
Interpretability refers to the extent to which a human can comprehend decisions \cite{Miller2019explanation}.
When systems consider individual differences, value-based explanations connect values to the reasoning behind decisions, making them more effective in justifying decisions \cite{Winikoff2021valuings,Tzeng2024ExannaA}.

Explainability techniques include rule-based approaches such as XCS \cite{ButzW2000XCS}, local interpretable model-agnostic explanations (LIME) method \cite{Ribeiro2016LIME}, and Shapley Additive explanations (SHAP) \cite{Lundberg2017SHAP}.
XCS combines reinforcement learning and genetic algorithms to learn the rules of decisions.
Specifically, rule-based approaches achieve interpretability by linking the decisive features to the decisions.
LIME approximates the behavior of the classifier for specific prediction and, therefore, enables interpretability. 
SHAP interprets the machine learning model by evaluating the contribution of each feature to the model's prediction.

The complexity, interdependence, and context-dependence of values significantly challenge explainability.
For instance, the trade-off between competing values is hard to measure and explain with contextual variability.
In addition, explanations typically include a premise and a consequent. The interdependence between values can lead to explanations that incorrectly attribute decisions to specific values.
Scalability extragates the difficulty of explainability when systems and environments become complex.
When systems lack full context, interpretations and explanations may become inaccurate and misleading.

\subsection*{Value Source}
\paragraph{Opportunity.}
Research can explore methods for continuously monitoring, updating, and adapting AI systems to maintain alignment with evolving values or intended values. Additionally, research can investigate approaches for aggregating and reconciling values at individual and group levels.

\paragraph{Challenge.}
The fundamental issue of value alignment is determining with whom and with what values the system should be aligned. The following list outlines the groups of individual values to consider.
\begin{itemize}
    \item \textbf{Direct users} of the systems
    \item \textbf{Decision-makers} responsible for deploying the system
    \item \textbf{Affected individuals} who are influenced by the system's decisions
    \item \textbf{Major population of society} that reflects the collective values
    \item \textbf{Vulnerable groups} whose values are necessary to avoid bias and harm
    \item \textbf{Policymakers} who decide what values to promote and demote
\end{itemize}

The following list highlights what values to consider.
\begin{itemize}
    \item \textbf{Ethical values} such as fairness, privacy, transparency, and accountability
    \item \textbf{Human-centered values} such as self-direction, achievement, or tradition
    \item \textbf{Localized values} such as norms in a group or culture
    \item \textbf{Domain-specific values} such as efficiency or sustainability
\end{itemize}

From a user's perspective, individuals may desire systems that offer suggestions consistent with both their personal values and broader societal values.
From a stakeholder's perspective, policymakers or decision-makers of organizations will expect systems to promote the values they choose.
The policymakers might aim to enhance fairness and privacy for social good while fostering transparency and accountability to build trust in systems.
Incorporating the values of vulnerable groups is necessary to enhance fairness and avoid bias.

The source of values introduces new challenges.
For instance, large language models become useful in varied domains and daily life.
However, the models reflect the predominant values present in the data used for training.
An individual from a culture with values different from the predominant one may find the suggestion ineffective.
Furthermore, different domains or cultures may interpret the same values in different ways.
These differences can lead to ineffective suggestions.

\subsection*{Applications}
\paragraph{Opportunity.}
Research can explore strategies for balancing agent autonomy with alignment. Additionally, research can investigate mechanisms that allow temporary deviations from alignment over extended periods to reflect real-world alignment.

\paragraph{Challenge.}
While the significance of value alignment in systems is increasingly recognized, it is not essential for every system.
For instance, robot vacuum cleaners can complete tasks without accounting for user values.
In essence, integrating values into systems is redundant in contexts that involve minimal ethical considerations or highly routine and predictable tasks, such as the operation of robotic arms in factories.
Further discussion is needed to determine the types of systems or applications in the real world where embedding values is essential.
For instance, incorporating fairness, privacy, and individual preferences into healthcare treatment recommendations is a clear and intuitive case.
Different values may drive individuals to choose different treatment strategies.
In addition, false positives or false negatives within different contexts involve different consequences of errors. 
Connecting these consequences to their respective values supports better decision making.
In the context of education, value alignment can enable personalized and inclusive learning by offering tailored feedback. Similarly, in social simulations, value alignment can assist policymakers in making decisions that reflect societal values or account for the values of different parties.
For autonomous vehicles or robotics, embedding ethical concerns can help to minimize the harm in accidents.

Another fundamental question is the balance between autonomy and alignment in multi-agent systems.
The ideal scenario is achieving high autonomy and high alignment, where agents can independently reason and act in complex environments while ensuring the decisions are consistent with values. The idea is that alignment with an abstract value leaves open many ways of how to fulfill a task, which gives an agent a high level of autonomy.
However, autonomy and alignment are sometimes at odds.
Strictly enforcing alignment constraints can limit agents' autonomy, while granting agents higher autonomy reduces human monitoring and intervention. Temporary deviations from value alignment can result in much better outcomes in the long run, but it becomes unclear when temporary deviations become permanent and should be corrected.
Less human control, however, can increase the risk of deviating from intended values.

Beyond the system design and implementation, a system designed for one society but deployed in another with different values can result in numerous challenges.
For instance, introducing a multi-agent system aligned with a culture of high individualism to a society with higher collectivism can create conflicts. The agents might consistently choose options that benefit themselves while trusting that the interaction mechanism takes care of a global optimum. However, this might lead to unacceptable solutions from a collectivistic point of view.
Such misalignments can lead to resistance against the systems.


\section{Conclusions}
Value alignment represents a non-trivial challenge in the development of responsible AI \cite{Dignum2019responsibleAI}.
AI systems must be aligned with human values to provide better support.
Addressing this challenge requires technical advancement and interdisciplinary collaboration to ensure AI systems respect and reflect the diverse values of humans.
In this paper, we argue that achieving value alignment requires a comprehensive theoretical foundation and should be modeled in terms of long-term dynamics within a multi-agent context.
While values in AI are often treated as static and universally shared concepts, real-world scenarios require a more nuanced understanding. In practice, values are context-dependent, held by individuals or groups, can evolve over time, and may vary or conflict within an individual and among individuals or groups. 
Furthermore, as AI systems increasingly operate in multi-agent environments, negotiating and coordinating for value differences and conflicts become inevitable.
Analyzing value alignment in multi-agent systems requires shifting perspectives between micro-level values and macro-level values, which refers to individual values and collective values.

The shifts we propose are essential and pragmatic. 
Without acknowledging the dynamic, long-term stable, and interdependent nature of values, AI systems risk reinforcing dominant values or short-term optimization, overlooking the complexity of values, and oversimplifying complex ethical and social considerations. This oversight can lead to alignment failures when AI systems operate in dynamic environments.
Research in multi-agent systems, with its emphasis on decentralized reasoning, emergent behavior, and coordination and competition, provides a natural and more effective frame for tackling these challenges that are beyond the capabilities of a single universal agent.
Lastly, it is important to acknowledge that value alignment is inherently an interdisciplinary challenge that computer scientists cannot define, identify, or implement values effectively without insights from other disciplines.

We envision value alignment as making decisions, assessing outcomes and situations, and providing justifications, all guided by underlying values.
Additionally, the systems must be capable of dynamically adapting to evolving values and contextual factors.
Achieving this vision requires research in several key areas, including eliciting values from human activities, embedding values into systems, verifying values against established requirements, validating values during real-world operation, interpreting system outputs related to values, explaining decisions based on values, and adapting to evolving values.
This vision opens up new research directions and encourages interdisciplinary collaboration to tackle one of the most pressing questions: How to develop AI that embodies human values and serves humanity.

\bibliography{SzTing,ref}

\begin{thebibliography}{33}
\providecommand{\natexlab}[1]{#1}
\providecommand{\url}[1]{\texttt{#1}}
\expandafter\ifx\csname urlstyle\endcsname\relax
  \providecommand{\doi}[1]{doi: #1}\else
  \providecommand{\doi}{doi: \begingroup \urlstyle{rm}\Url}\fi

\bibitem[Ajmeri et~al.(2020)Ajmeri, Guo, Murukannaiah, and Singh]{Ajmeri2020elessar}
N.~Ajmeri, H.~Guo, P.~K. Murukannaiah, and M.~P. Singh.
\newblock Elessar: Ethics in norm-aware agents.
\newblock In \emph{Proceedings of the 19th International Conference on Autonomous Agents and Multiagent Systems, ({AAMAS})}, pages 16--24, Auckland, May 2020. IFAAMAS.
\newblock \doi{10.5555/3398761.3398769}.

\bibitem[Brown et~al.(2021)Brown, Schneider, Dragan, and Niekum]{Brown2021VAV}
D.~S. Brown, J.~Schneider, A.~Dragan, and S.~Niekum.
\newblock Value alignment verification.
\newblock In \emph{Proceedings of the 38th International Conference on Machine Learning}, volume 139 of \emph{Proceedings of Machine Learning Research}, pages 1105--1115, Virtual, Jul 2021. PMLR, PMLR.

\bibitem[Butz and Wilson(2000)]{ButzW2000XCS}
M.~V. Butz and S.~W. Wilson.
\newblock An algorithmic description of {XCS}.
\newblock In \emph{Proceedings of the 3rd International Workshop on Learning Classifier Systems}, volume 1996 of \emph{LNCS}, pages 253--272, Paris, France, 2000. Springer.
\newblock \doi{10.1007/3-540-44640-0_15}.

\bibitem[Chakraborty et~al.(2024)Chakraborty, Bedi, Koppel, Wang, Manocha, Wang, and Huang]{Chakraborty2024PARL}
S.~Chakraborty, A.~Bedi, A.~Koppel, H.~Wang, D.~Manocha, M.~Wang, and F.~Huang.
\newblock Parl: {A} unified framework for policy alignment in reinforcement learning from human feedback.
\newblock In \emph{Proceedings of the 12th International Conference on Learning Representations (ICLR)}, pages 1--40, Vienna, May 2024. ICLR.

\bibitem[Chopra and Singh(2018)]{Chopra2018ethics}
A.~K. Chopra and M.~P. Singh.
\newblock Sociotechnical systems and ethics in the large.
\newblock In \emph{Proceedings of the AAAI/ACM Conference on Artificial Intelligence, Ethics, and Society (AIES)}, pages 48--53, New Orleans, Feb. 2018. ACM.
\newblock \doi{10.1145/3278721.3278740}.

\bibitem[Dennis et~al.(2016)Dennis, Fisher, Slavkovik, and Webster]{Dennis2016formal}
L.~Dennis, M.~Fisher, M.~Slavkovik, and M.~Webster.
\newblock Formal verification of ethical choices in autonomous systems.
\newblock \emph{Robotics and Autonomous Systems}, 77:\penalty0 1--14, 2016.
\newblock \doi{10.1016/j.robot.2015.11.012}.

\bibitem[Di~Scala and Yolum(2023)]{Di2023paccart}
D.~Di~Scala and P.~Yolum.
\newblock Paccart: Reinforcing trust in multiuser privacy agreement systems.
\newblock In \emph{Proceedings of the 22nd International Conference on Autonomous Agents and Multiagent Systems ({AAMAS})}, pages 2787--2789, London, 2023. IFAAMAS.
\newblock \doi{10.5555/3545946.3599078}.

\bibitem[Dignum(2019)]{Dignum2019responsibleAI}
V.~Dignum.
\newblock \emph{Responsible Artificial Intelligence: How to Develop and Use AI in A Responsible Way}.
\newblock Artificial Intelligence: Foundations, Theory, and Algorithms. Springer, Cham, 2019.
\newblock \doi{10.1007/978-3-030-30371-6}.

\bibitem[Friedman et~al.(2013)Friedman, Kahn, Borning, and Huldtgren]{Friedman2013VSD}
B.~Friedman, P.~H. Kahn, A.~Borning, and A.~Huldtgren.
\newblock Value sensitive design and information systems.
\newblock \emph{Early Engagement and New Technologies: {Opening} Up the Laboratory}, 16:\penalty0 55--95, 2013.
\newblock \doi{10.1007/978-94-007-7844-3_4}.

\bibitem[Gabriel(2020)]{Gabriel2020AI}
I.~Gabriel.
\newblock Artificial intelligence, values, and alignment.
\newblock \emph{Minds and machines}, 30\penalty0 (3):\penalty0 411--437, 2020.
\newblock \doi{10.1007/s11023-020-09539-2}.

\bibitem[Haidt et~al.(2009)Haidt, Graham, and Joseph]{Haidt2009aboveBelow}
J.~Haidt, J.~Graham, and C.~Joseph.
\newblock Above and below left--right: Ideological narratives and moral foundations.
\newblock \emph{Psychological Inquiry}, 20\penalty0 (2-3):\penalty0 110--119, 2009.
\newblock \doi{10.1080/10478400903028573}.

\bibitem[Lera-Leri et~al.(2022)Lera-Leri, Bistaffa, Serramia, Lopez-Sanchez, and Rodriguez-Aguilar]{Lera2022towards}
R.~Lera-Leri, F.~Bistaffa, M.~Serramia, M.~Lopez-Sanchez, and J.~Rodriguez-Aguilar.
\newblock Towards pluralistic value alignment: Aggregating value systems through $\ell$p-regression.
\newblock In \emph{Proceedings of the 21st International Conference on Autonomous Agents and Multiagent Systems ({AAMAS})}, pages 780--788, Auckland, 2022. IFAAMAS.
\newblock \doi{10.5555/3535850.3535938}.

\bibitem[Liscio et~al.(2021)Liscio, van~der Meer, Siebert, Jonker, Mouter, and Murukannaiah]{Liscio2021axies}
E.~Liscio, M.~van~der Meer, L.~C. Siebert, C.~M. Jonker, N.~Mouter, and P.~K. Murukannaiah.
\newblock Axies: Identifying and evaluating context-specific values.
\newblock In \emph{Proceedings of the 20th International Conference on Autonomous Agents and MultiAgent Systems ({AAMAS})}, pages 799--808, London, 2021. IFAAMAS.
\newblock \doi{10.5555/3463952.3464048}.

\bibitem[Liscio et~al.(2023)Liscio, Lera-Leri, Bistaffa, Dobbe, Jonker, Lopez-Sanchez, Rodriguez-Aguilar, and Murukannaiah]{Liscio2023value}
E.~Liscio, R.~Lera-Leri, F.~Bistaffa, R.~I. Dobbe, C.~M. Jonker, M.~Lopez-Sanchez, J.~A. Rodriguez-Aguilar, and P.~K. Murukannaiah.
\newblock Value inference in sociotechnical systems.
\newblock In \emph{Proceedings of the 22nd International Conference on Autonomous Agents and Multiagent Systems ({AAMAS})}, pages 1774--1780, London, 2023. IFAAMAS.
\newblock \doi{10.5555/3545946.3598838}.
\newblock {Blue} {Sky} {Ideas} {Track}.

\bibitem[Lundberg and Lee(2017)]{Lundberg2017SHAP}
S.~M. Lundberg and S.-I. Lee.
\newblock A unified approach to interpreting model predictions.
\newblock In \emph{Proceedings of the 31st International Conference on Neural Information Processing Systems (NIPS)}, pages 4768--4777, Long Beach, 2017. Curran Associates.
\newblock \doi{10.5555/3295222.3295230}.

\bibitem[Mechergui and Sreedharan(2024)]{Mechergui2024goalAlignment}
M.~Mechergui and S.~Sreedharan.
\newblock Goal alignment: {Re}-analyzing value alignment problems using human-aware ai.
\newblock In \emph{Proceedings of the 38th AAAI Conference on Artificial Intelligence}, volume~38, pages 10110--10118, Vancouver, 2024. {AAAI} Press.
\newblock \doi{10.1609/aaai.v38i9.28875}.

\bibitem[Miller(2019)]{Miller2019explanation}
T.~Miller.
\newblock Explanation in artificial intelligence: Insights from the social sciences.
\newblock \emph{Artificial Intelligence}, 267:\penalty0 1--38, 2019.
\newblock \doi{10.1016/j.artint.2018.07.007}.

\bibitem[Murukannaiah et~al.(2020)Murukannaiah, Ajmeri, Jonker, and Singh]{Murukannaiah2020Blue-Sky}
P.~K. Murukannaiah, N.~Ajmeri, C.~M. Jonker, and M.~P. Singh.
\newblock New foundations of ethical multiagent systems.
\newblock In \emph{Proceedings of the 19th International Conference on Autonomous Agents and Multiagent Systems ({AAMAS})}, pages 1706--1710, Auckland, May 2020. IFAAMAS.
\newblock \doi{10.5555/3398761.3398958}.
\newblock {Blue} {Sky} {Ideas} {Track}.

\bibitem[Osman and d’Inverno(2024)]{Osman2024values}
N.~Osman and M.~d’Inverno.
\newblock A computational framework of human values.
\newblock In \emph{Proceedings of the 23rd Conference on Autonomous Agents and MultiAgent Systems (AAMAS)}, pages 1531--1539, Auckland, May 2024. IFAAMAS.
\newblock \doi{10.5555/3635637.3663013}.

\bibitem[Ribeiro et~al.(2016)Ribeiro, Singh, and Guestrin]{Ribeiro2016LIME}
M.~T. Ribeiro, S.~Singh, and C.~Guestrin.
\newblock "why should i trust you?" {Explaining} the predictions of any classifier.
\newblock In \emph{Proceedings of the 22nd ACM SIGKDD international conference on knowledge discovery and data mining}, pages 1135--1144, San Francisco, 2016. ACM.
\newblock \doi{10.1145/2939672.293977}.

\bibitem[Rodriguez-Soto et~al.(2023)Rodriguez-Soto, Radulescu, Rodriguez-Aguilar, Lopez-Sanchez, and Now{\'e}]{Rodriguez2023MORL}
M.~Rodriguez-Soto, R.~Radulescu, J.~A. Rodriguez-Aguilar, M.~Lopez-Sanchez, and A.~Now{\'e}.
\newblock Multi-objective reinforcement learning for guaranteeing alignment with multiple values.
\newblock In \emph{Proceedings of the Adaptive and Learning Agents Workshop (ALA)}, LNCS, pages 1--9, London, May 2023. Springer.

\bibitem[Rokeach(1973)]{Rokeach1973values}
M.~Rokeach.
\newblock \emph{The Nature of Human Values}.
\newblock Free Press, New York, 1973.

\bibitem[Russell(2019)]{Russell2019human}
S.~Russell.
\newblock \emph{Human Compatible: {AI} and the Problem of Control}.
\newblock Penguin, New York, 2019.

\bibitem[Schwartz(1994)]{Schwartz1994UniversalAspectsOfValues}
S.~H. Schwartz.
\newblock Are there universal aspects in the structure and contents of human values?
\newblock \emph{Journal of Social Issues}, 50\penalty0 (4):\penalty0 19--45, 1994.
\newblock \doi{10.1111/j.1540-4560.1994.tb01196.x}.

\bibitem[Schwartz(2012)]{Schwartz2012overview}
S.~H. Schwartz.
\newblock An overview of the {Schwartz} theory of basic values.
\newblock \emph{Online Readings in Psychology and Culture}, 2\penalty0 (1):\penalty0 3--20, 2012.
\newblock \doi{10.9707/2307-0919.1116}.

\bibitem[Serramia et~al.(2018)Serramia, L{\'{o}}pez{-}S{\'{a}}nchez, Rodr{\'{\i}}guez{-}Aguilar, Rodr{\'{\i}}guez, Wooldridge, Morales, and Ans{\'{o}}tegui]{Serramia2018values}
M.~Serramia, M.~L{\'{o}}pez{-}S{\'{a}}nchez, J.~A. Rodr{\'{\i}}guez{-}Aguilar, M.~Rodr{\'{\i}}guez, M.~Wooldridge, J.~Morales, and C.~Ans{\'{o}}tegui.
\newblock Moral values in norm decision making.
\newblock In \emph{Proceedings of the 17th Conference on Autonomous Agents and MultiAgent Systems (AAMAS)}, pages 1294--1302, Stockholm, July 2018. IFAAMAS.
\newblock \doi{10.5555/3237383.3237891}.

\bibitem[Serramia et~al.(2023)Serramia, Rodriguez-Soto, Lopez-Sanchez, Rodriguez-Aguilar, Bistaffa, Boddington, Wooldridge, and Ansotegui]{Serramia2023encoding}
M.~Serramia, M.~Rodriguez-Soto, M.~Lopez-Sanchez, J.~A. Rodriguez-Aguilar, F.~Bistaffa, P.~Boddington, M.~Wooldridge, and C.~Ansotegui.
\newblock Encoding ethics to compute value-aligned norms.
\newblock \emph{Minds and Machines}, 33\penalty0 (4):\penalty0 761--790, 2023.
\newblock \doi{10.1007/s11023-023-09649-7}.

\bibitem[Shen et~al.(2024)Shen, Knearem, Ghosh, Alkiek, Krishna, Liu, Ma, Petridis, Peng, Qiwei, Rakshit, Si, Xie, Bigham, Bentley, Chai, Lipton, Mei, Mihalcea, Terry, Yang, Morris, Resnick, and Jurgens]{Shen2024bidirectionalAlignment}
H.~Shen, T.~Knearem, R.~Ghosh, K.~Alkiek, K.~Krishna, Y.~Liu, Z.~Ma, S.~Petridis, Y.-H. Peng, L.~Qiwei, S.~Rakshit, C.~Si, Y.~Xie, J.~P. Bigham, F.~Bentley, J.~Chai, Z.~Lipton, Q.~Mei, R.~Mihalcea, M.~Terry, D.~Yang, M.~R. Morris, P.~Resnick, and D.~Jurgens.
\newblock Towards bidirectional human-{AI} alignment: {A} systematic review for clarifications, framework, and future directions.
\newblock \emph{arXiv preprint arXiv:2406.09264}, 2406.09264:\penalty0 1--56, 2024.

\bibitem[Sierra et~al.(2021)Sierra, Osman, Noriega, Sabater-Mir, and Perell{\'o}]{Sierra2021formalVA}
C.~Sierra, N.~Osman, P.~Noriega, J.~Sabater-Mir, and A.~Perell{\'o}.
\newblock Value alignment: {A} formal approach.
\newblock \emph{arXiv preprint arXiv:2110.09240}, 2110.09240:\penalty0 1--15, 2021.

\bibitem[Sorensen et~al.(2024)Sorensen, Jiang, Hwang, Levine, Pyatkin, West, Dziri, Lu, Rao, Bhagavatula, et~al.]{Sorensen2024value}
T.~Sorensen, L.~Jiang, J.~D. Hwang, S.~Levine, V.~Pyatkin, P.~West, N.~Dziri, X.~Lu, K.~Rao, C.~Bhagavatula, et~al.
\newblock Value kaleidoscope: Engaging {AI} with pluralistic human values, rights, and duties.
\newblock In \emph{Proceedings of the AAAI Conference on Artificial Intelligence}, volume~38, pages 19937--19947. {AAAI} Press, 2024.
\newblock \doi{10.1609/aaai.v38i18.29970}.

\bibitem[Tzeng et~al.(2024)Tzeng, Ajmeri, and Singh]{Tzeng2024ExannaA}
S.-T. Tzeng, N.~Ajmeri, and M.~P. Singh.
\newblock Value-based rationales improve social experience: {A} multiagent simulation study.
\newblock \emph{arXiv preprint arXiv:2408.02117}, 2408.02117:\penalty0 1--13, 2024.

\bibitem[Winikoff et~al.(2021)Winikoff, Sidorenko, Dignum, and Dignum]{Winikoff2021valuings}
M.~Winikoff, G.~Sidorenko, V.~Dignum, and F.~Dignum.
\newblock Why bad coffee? {Explaining} {BDI} agent behaviour with valuings.
\newblock \emph{Artificial Intelligence}, 300:\penalty0 103554, 2021.
\newblock \doi{10.1016/j.artint.2021.103554}.

\bibitem[Woodgate and Ajmeri(2022)]{Woodgate2022Blue-Sky}
J.~Woodgate and N.~Ajmeri.
\newblock Macro ethics for governing equitable sociotechnical systems.
\newblock In \emph{Proceedings of the 21st International Conference on Autonomous Agents and Multiagent Systems (AAMAS)}, pages 1824--1828, Online, May 2022. IFAAMAS.
\newblock \doi{10.5555/3535850.3536118}.
\newblock {Blue} {Sky} {Ideas} {Track}.

\end{thebibliography}
\end{document}